\documentclass[letterpaper, 10 pt, conference]{ieeeconf}  

\IEEEoverridecommandlockouts                              

\usepackage{amsmath}
\usepackage{amssymb}
\usepackage{bm}
\usepackage{graphicx}
\usepackage{multirow}
\usepackage{subcaption}
\usepackage{svg}
\usepackage{xfrac}
\usepackage{xparse}

\NewDocumentCommand{\scnum}{ >{\SplitArgument{1}{e}}m }
 {\scnumaux#1}
\NewDocumentCommand{\scnumaux}{ m m }
 {#1\,\mathrm{e}{#2}}

\title{\LARGE \bf
RV-FuseNet: Range View Based Fusion of Time-Series LiDAR Data for Joint 3D Object Detection and Motion Forecasting
}

\author{Ankit Laddha$^{*1,2}$, Shivam Gautam$^{*1,2}$, Gregory P. Meyer$^{1,3}$, Carlos Vallespi-Gonzalez$^{1,2}$ and Carl K. Wellington$^{1,2}$
\thanks{*Equal Contribution}

\thanks{$^{1}$ Work done while at Uber Advanced Technologies Group, Pittsburgh}
\thanks{$^{2}$ Aurora Innovation, Pittsburgh, USA}
\thanks{$^{3}$ Motional, Pittsburgh, USA}
}

\begin{document}

\maketitle
\thispagestyle{empty}
\pagestyle{empty}

\begin{abstract}

Robust real-time detection and motion forecasting of traffic participants is necessary for autonomous vehicles to safely navigate urban environments. In this paper, we present RV-FuseNet, a novel end-to-end approach for joint detection and trajectory estimation directly from time-series LiDAR data. Instead of the widely used bird's eye view (BEV) representation, we utilize the native range view (RV) representation of LiDAR data. The RV preserves the full resolution of the sensor by avoiding the voxelization used in the BEV. Furthermore, RV can be processed efficiently due to its compactness. Previous approaches project time-series data to a common viewpoint for temporal fusion, and often this viewpoint is different from where it was captured. This is sufficient for BEV methods, but for RV methods, this can lead to loss of information and data distortion which has an adverse impact on performance. To address this challenge we propose a simple yet effective novel architecture, \textit{Incremental Fusion}, that minimizes the information loss by sequentially projecting each RV sweep into the viewpoint of the next sweep in time. We show that our approach significantly improves motion forecasting performance over the existing state-of-the-art. Furthermore, we demonstrate that our sequential fusion approach is superior to alternative RV based fusion methods on multiple datasets.

\end{abstract}

\section{Introduction}
\label{sec:intro}

Autonomous vehicles need to perform path planning in dense and dynamic urban environments where many actors are simultaneously trying to navigate. This requires robust and efficient methods for object detection and motion forecasting. The goal of detection is to recognize the objects present in the scene and estimate their 3D bounding box, while prediction aims to estimate the position of the detected boxes in the future. Traditionally, separate models have been used for detection and prediction. Recently, joint methods \cite{spagnn, faf, intentnet, laserflow, liranet} using LiDAR have been proposed. These methods benefit from reduced latency, feature sharing across both tasks and lower computational requirements for detection and prediction.

\begin{figure}[t]
    \centering
    \includegraphics[width=0.48\textwidth]{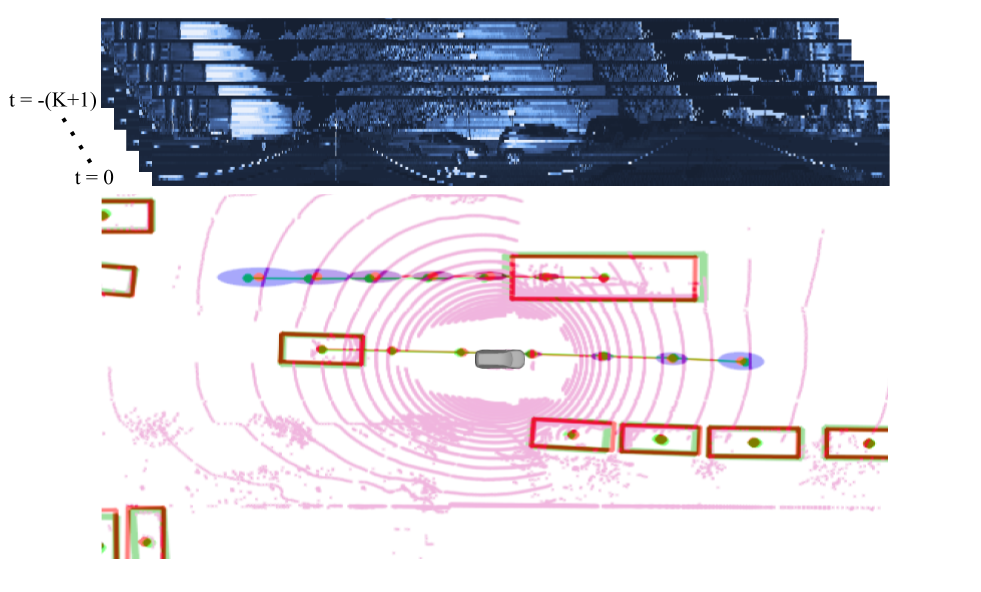}
\caption{\textbf{Top:} Input time-series LiDAR sweeps in the native range view . \textbf{Bottom:} Output 3D detections and predicted trajectories with position uncertainties (ground truth in green).}
\label{fig:task}
\vspace{-1em}
\end{figure}

As shown in Fig.~\ref{fig:task}, the input data captured from a LiDAR sensor is natively in the form of a range view image.
Each sweep of the LiDAR only provides measurements of the instantaneous position of objects. Therefore, multiple sweeps are required for motion forecasting.
The input LiDAR data is generally organized in one of three different ways, depending on how much pre-processing is performed; the raw range view (RV) measurements from the sensor~\cite{laserflow, lasernet}, a 3D point cloud created from the range measurements~\cite{pointrcnn, flownet3D}, or a 2D/3D bird's eye view (BEV) image~\cite{intentnet,pixor} which computes summary features of the points in a voxelized grid.

We desire to preserve the full resolution of the raw sensor data, but this can result in computational challenges. For example, one approach for processing a time-series of 3D points is by creating a k-nearest neighbor graph between the points in subsequent frames~\cite{flownet3D}, but this can be prohibitively expensive with tens or hundreds of thousands of LiDAR points as is common in autonomous driving. Therefore, most of the end-to-end methods~\cite{faf, intentnet, stinet} use a BEV grid to represent and aggregate the time-series LiDAR data. These methods compute the features of each cell using the LiDAR points falling in the cell, and then use a convolutional neural network (CNN) to learn the motion of BEV cells across time. This pre-processing step makes them faster than the raw point-based methods, but the computational gain comes with the loss of high resolution information due to discretization. In comparison, the RV representation~\cite{lasernet} preserves the high resolution point information while maintaining a compact representation that can be processed efficiently. Additionally, the RV preserves the occlusion information across objects in the scene that can be lost when the original range measurements are projected into BEV. However, the RV introduces significant complexity when combining a time-series of input sensor data.

\begin{figure}[t]
\centering
\begin{subfigure}{0.23\textwidth}
\centering
\includegraphics[width=\textwidth]{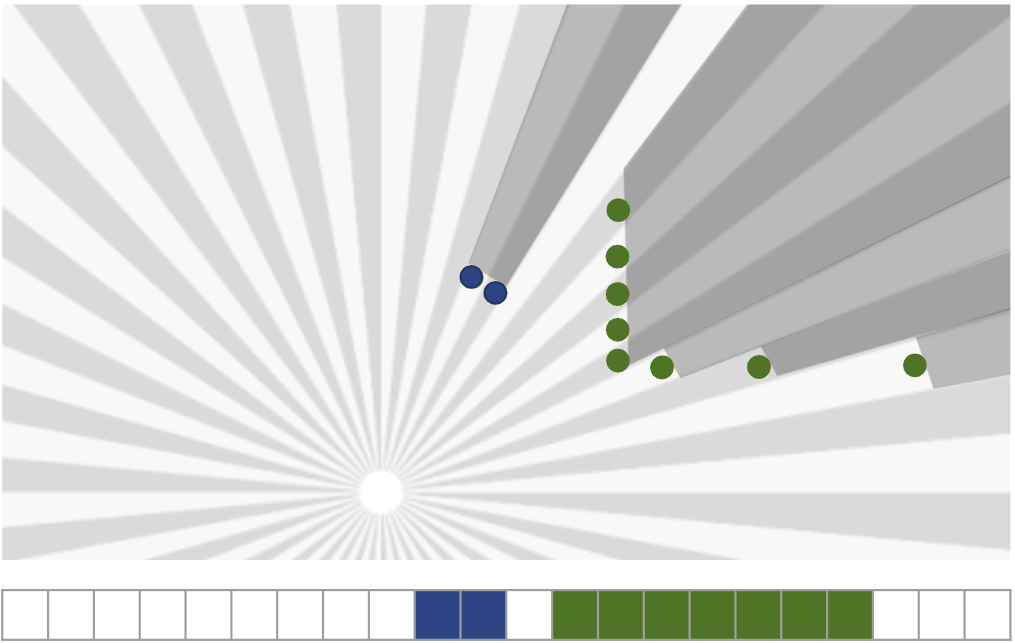}
\label{fig:before_distortion}
\end{subfigure}
\hspace{0.5em}
\begin{subfigure}{0.23\textwidth}
\centering
\includegraphics[width=\textwidth]{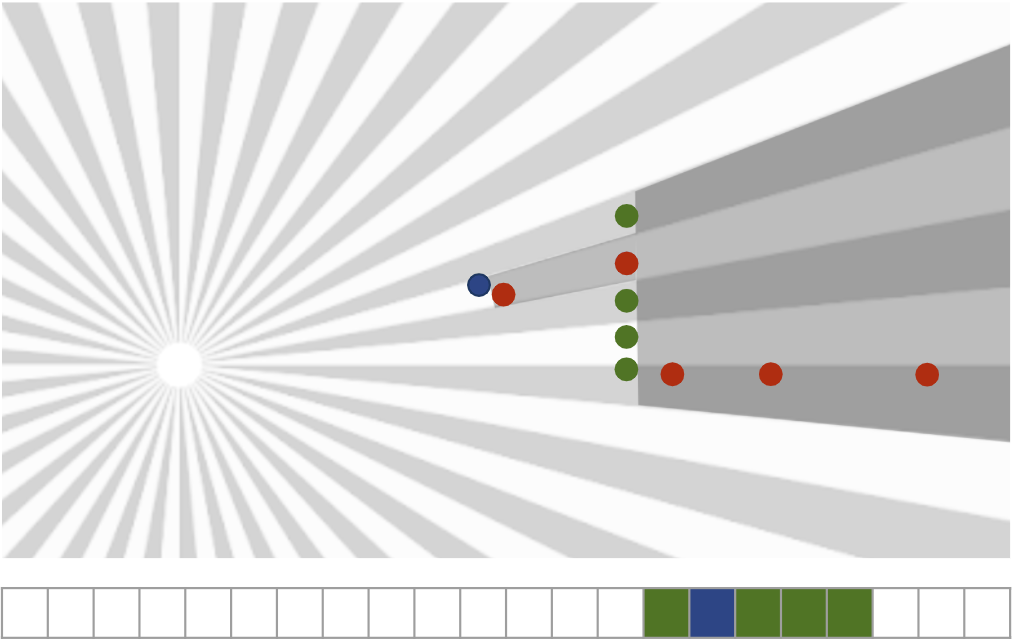}
\label{fig:after_distortions}
\end{subfigure}
\vspace{-1em}
\caption{\textbf{Range View Challenges:} We depict a scene with a LiDAR sweep captured from a single laser beam in two different viewpoints. We show the range view angular bins (top) as well as the array bin representing the rasterized image input to the model (bottom). On the left, we show the sweep from its original viewpoint where the data was captured. The measurements for the two objects in the scene are shown in blue and green, respectively, and the occluded area behind them is shaded. In this case, each angular bin contains at most one LiDAR point, thus preserving all the points captured by the sensor. On the right, we show the same LiDAR points from a different viewpoint. We see some points (depicted in red) are lost due to occlusions from other objects, self-occlusions, or due to multiple points falling in the same bin. It is noteworthy that the two separate objects appear intermingled in the discretized array bin. }
\label{fig:distortion}
\vspace{-1em}
\end{figure}

Existing end-to-end methods \cite{spagnn, faf, intentnet, laserflow} assume ego-motion is estimated separately using localization techniques, and we follow the same setup. This enables decoupling object motion and ego-motion to reduce the complexity of the problem.
Ego-motion compensation is achieved by changing the viewpoint of past data from the original ego-frame to the current one. In the BEV, this compensation requires only translation and rotation which preserves the original data without any additional information loss. However, in the RV this change in viewpoint requires a projection that can create gaps in continuous surfaces, as well as information loss due to self-occlusions, occlusions between objects and discretization effects (see Fig.~\ref{fig:distortion}). Furthermore, the magnitude of these effects is dependent on the amount of ego-motion, the speed of the objects, and the time difference between the viewpoints. This makes time-series fusion of LiDAR data in the RV challenging. Minimizing this information loss and its impact is critical for good performance. Recently, another RV based method~\cite{laserflow}, proposed to pre-process each sweep in it's own frame before fusion to remedy the information loss. However, the temporal fusion is performed in the most recent viewpoint, which still leads to significant information loss.

In this work, we propose RV-FuseNet, a novel end-to-end joint detection and motion forecasting method using the native RV representation of LiDAR data. As discussed previously, RV has the benefit of maintaining the high resolution point information, but comes with the challenges of projection change and data loss due to changing viewpoints. We propose a novel fusion architecture, termed \textit{Incremental Fusion}, to tackle these challenges. It sequentially fuses the temporal LiDAR data such that the data collected from one viewpoint is only projected to the nearest viewpoint in the future, so as to minimize the information loss. We experiment with two LiDARs with different resolutions to show that this architecture is highly effective and outperforms other fusion techniques. We further demonstrate that it is particularly effective in cases where high-speed ego or object motion is present. Finally, using the proposed \textit{Incremental Fusion} architecture, we establish a new state-of-the-art result for detection and motion forecasting performance on the publicly available  nuScenes~\cite{nuscenes} dataset, by showing a ${\sim17\%}$ improvement on trajectory prediction over the previous best method.

\begin{figure*}[!ht]
    \centering
    \includegraphics[width=0.82\textwidth]{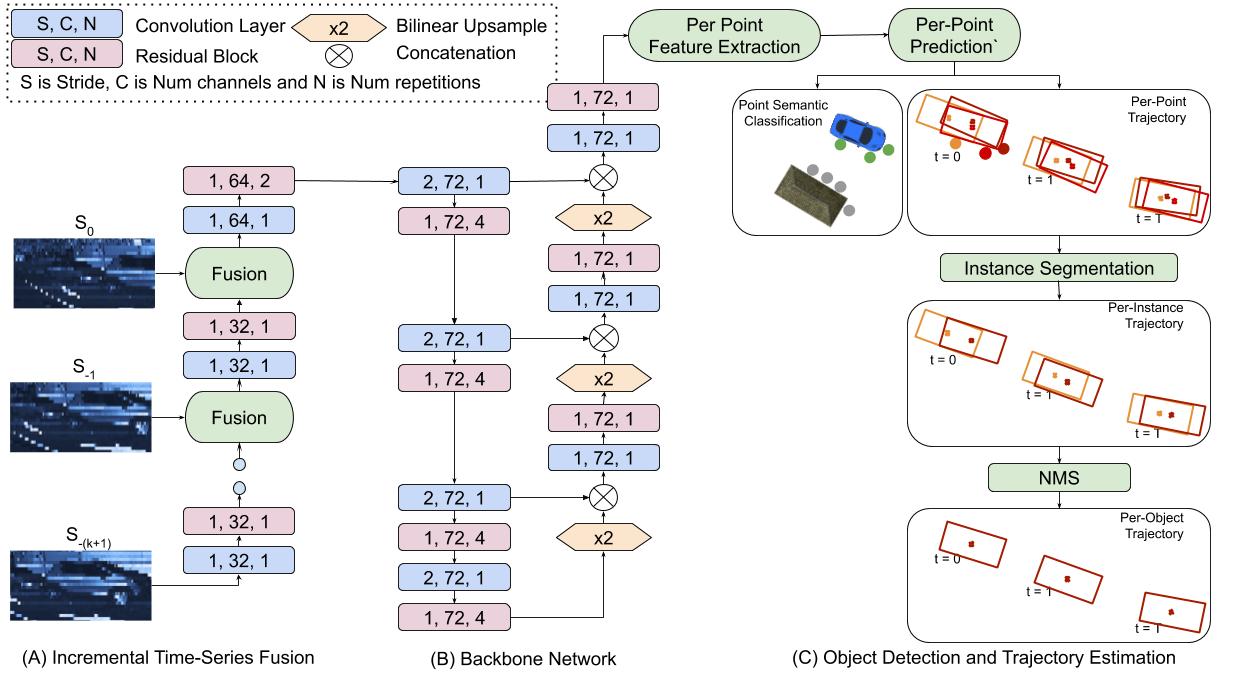}
    \caption{\textbf{RV-FuseNet Overview:} We propose an incremental scheme (A) which sequentially fuses a time-series of LiDAR sweeps in the RV for extracting spatio-temporal features. The \textit{Fusion} of consecutive sweeps is done by warping the previous sweep features into the viewpoint of the next sweep and concatenating them with the next sweep's features (Section~\ref{sec:incremenal_fusion} and~\ref{sec:rv_ts_fusion}). The final spatio-temporal features are further processed using (B) the backbone to generate features for each LiDAR point in the most recent sweep. These per-point features are used to predict (C) the semantic class and trajectory of each point. The points are segmented into instances, and a trajectory for each instance is computed by averaging the per-point trajectories. Finally, NMS is applied on the boxes at $t=0$ to remove duplicate detections and their trajectories.}
    \label{fig:overview}
\vspace{-1em}
\end{figure*}

\section{Related Work}

\label{sec:related_work}
\textbf{Multi-Stage Forecasting:} Fusion of temporal sensor data is of paramount importance for robust motion estimation. The most popular methods of enforcing temporal consistency involve using detection-based approaches~\cite{lasernet, pointrcnn, pixor, lasernet++, pointpillars, zhouVoxelNetEndtoEndLearning2018, ngiam2019starnet, multiview, pillarbaseddetection, centerdet, std} to generate objects. These detections are stitched across time through tracking \cite{milan2017online, sadeghian2017tracking, schulter2017deep, gautam2020sdvtracker} to estimate the state (position, velocity and acceleration) using a motion model. This state is further extrapolated into the future using the motion model for forecasting. These detect-then-track methods have good short term (i.e. $<1s$) forecasting performance but are not suitable for long-term trajectory predictions that are needed to navigate dense, urban environments~\cite{dp}. 

To overcome the shortcomings of direct state extrapolation methods for long term predictions,~\cite{dp, cui2019multimodal,rulesofroad,covernet} proposed to use learning based methods using the associated sequence of detections along with state information. Other approaches~\cite{sociallstm, desire, convolutionalsocialpooling, carnet, malla2020titan,multipath,lanegcn,destination,mfp} ignore the state and only use the associated temporal object detections as input to recurrent networks. However, such approaches rely on the output of detection and temporal association systems, making them susceptible to any changes and errors in the upstream models~\cite{spagnn}. For practical applications, this often results in cascaded pipelines which are hard to train and maintain. Further, such approaches lose out on rich feature representations learned by object detection networks.

\textbf{Future Occupancy Prediction:} Another class of methods for forecasting focuses on learning a probability distribution over the future Occupancy of Grid Maps (OGMs)~\cite{Mohajerin_2019, Hoermann_2018, Schreiber_2019}. They replace the multi-stage systems with a single model to directly predict grid occupancy from sensor data. However, these approaches fail to model actor-specific characteristics like class attributes and kinematic constraints. Furthermore, modeling complex interactions among actors and maintaining a balance between discretization resolution and computational efficiency remains a challenge~\cite{5980084}.

\textbf{End-to-End Detection and Prediction:} In contrast to multi-stage and OGM based methods, end-to-end approaches jointly learn object detection and motion forecasting \cite{spagnn, faf, intentnet, laserflow}. This reduces latency and system complexity by eliminating multiple stages while still preserving actor specific characteristics of the prediction model. In this work, we also present an end-to-end method. In the seminal work,~\cite{faf} proposed to jointly solve detection and prediction using a BEV representation of LiDAR. This was improved in~\cite{intentnet} by using maps and further improved in~\cite{spagnn} by using a two-stage model to incorporate interaction among objects. In contrast, we use the RV and show that by preserving the high resolution sensor information, we are able to outperform BEV methods by a significant margin with just a single stage.

Recently,~\cite{laserflow} also proposed an end-to-end method using a RV representation of LiDAR and showed that RV based method can also achieve similar results as BEV based methods. However, similar to BEV based methods,~\cite{laserflow} projects all temporal data into the most recent viewpoint which leads to large information loss from the past sweeps.
Unlike the one-shot temporal fusion approach of \cite{laserflow}, we propose sequential fusion of time-series data and show improved results.

\section{Range View based Detection and Prediction}
\label{sec:approach}
Fig.~\ref{fig:overview} shows an overview of our proposed approach for joint object detection and motion forecasting directly from sensor data. As the LiDAR sensor rotates, it continuously generates measurements. This data is sliced into ``sweeps'', wherein each slice contains measurements from a full $360^\circ$ rotation. We extract per-point spatio-temporal features using a time-series of sweeps as defined in Section~\ref{sec:input}. We use the native range view (RV) representation discussed in Section~\ref{sec:projection} for feature learning. To tackle the challenges of temporal fusion in the RV, we propose a novel scheme for sequentially combining the time-series of data in Section~\ref{sec:incremenal_fusion} leveraging the technique described in Section~\ref{sec:rv_ts_fusion}. Finally, in Section~\ref{sec:targets}, we describe how per-point features are used to generate the final detections and their trajectories. We train our method end-to-end using a probabilistic loss function, as  discussed in Section~\ref{sec:loss}.

\subsection{Input Time-Series}
\label{sec:input}
We assume that we are given a time-series of $K$ sweeps, denoted by $\{\mathcal{S}_{k}\}_{k=0}^{1-K}$, where $k=0$ is the most recent sweep and $-K < k < 0$ are the past sweeps. Our goal is to detect objects observed in the current sweep $\mathcal{S}_{0}$ and predict their trajectory. Each LiDAR sweep contains $N_{k}$ range measurements, which can be transformed into a set of 3D points, $\mathcal{S}_{k}= \{\bm{p}^{i}_{k} \}_{i = 1}^{N_k}$, using the provided pose (viewpoint) of the sensor $\mathcal{P}_{k}$ when the sweep was captured. Since the pose at each sweep is known, we can calculate the transformation of points from one viewpoint to another. We denote the $m$-th sweep transformed into the $n$-th sweep's coordinate frame as, $\mathcal{S}_{m,n} = \{\bm{p}^{i}_{m,n}\}_{i = 1}^{N_m}$, where each point is represented by its 3D coordinates, $[x^{i}_{m,n}, y^{i}_{m,n}, z^{i}_{m,n}]^T$.

For each point $\bm{p}^{i}_{m}$, we define a set of associated features $\bm{f}^{i}_{m}$ as: its range $r^{i}_{m}$ and azimuth $\theta^{i}_{m}$ in the frame it was captured, the remission or intensity $e^{i}_{m}$ of the LiDAR return, and its range $r^{i}_{m, 0}$ and azimuth $\theta^{i}_{m, 0}$ in the most recent frame.

\subsection{Range View Projection}
\label{sec:projection}
For a given sweep $\mathcal{S}_{m}$, we form a range view image using viewpoint $\mathcal{P}_{n}$, by defining the projection of a point $\bm{p}^{i}_{m,n}$ as,
\begin{equation}
\label{eq:projection}
\text{P}(\bm{p}^{i}_{m,n}) = \left(\left\lfloor \Phi^{i}_{m,n} / \Delta \Phi \right\rfloor, \left\lfloor \theta^{i}_{m,n} / \Delta \theta \right\rfloor\right)
\end{equation}
where the row of the range image is specified by the discretized elevation $\Phi^{i}_{m,n}$ and similarly, the column is defined by the discretized azimuth $\theta^{i}_{m,n}$. The angular resolutions of the image $\Delta \Phi$ and $\Delta \theta$ are dependent on the specific LiDAR structure. Furthermore, if more than one point projects into the same image coordinate, we keep the point with the smallest range, $r^{i}_{m,n}$. By applying $\text{P}$ to every point in $\mathcal{S}_{m,n}$, we can generate a range view image $\mathcal{I}_{m,n}$. Note that every sweep $\mathcal{S}_{k}$ projected in its original viewpoint $\mathcal{P}_{k}$ results in the native range image produced by LiDAR.

\subsection{Incremental Fusion}
\label{sec:incremenal_fusion}

The goal of fusion is to extract spatio-temporal features from a time-series of LiDAR sweeps. The most straightforward strategy used by many previous BEV based methods~\cite{spagnn, intentnet,pointpillars} is to \textit{warp} all the past sweeps into the most recent viewpoint and fuse them using a CNN. We refer to this strategy as \textit{Early Fusion}. However, in contrast to BEV, RV projection can lose information and generate gaps when the viewpoints are different (see Fig.~\ref{fig:distortion}). These effects are exacerbated as the distance between viewpoints increases. Therefore, \cite{laserflow} proposed \textit{Late Fusion} which first pre-processes each sweep in it's own viewpoint. It then performs the temporal fusion by projecting all the sweeps into the range view of the most recent sweep. However, since the fusion is done in a single viewpoint, there is significant information loss. Therefore, in this work we propose \textit{Incremental Fusion}  (Fig.~\ref{fig:overview}A), which minimizes the distance  between viewpoints during fusion.

Our \textit{Incremental Fusion} strategy sequentially fuses the sweeps from one timestep to the next. For each sweep, we extract spatio-temporal features in its original viewpoint by combining the features from that sweep with the spatio-temporal features from the past sweep. The  previous sweep is warped and merged with the current sweep using the method described in the Section \ref{sec:rv_ts_fusion}. The gradual change in viewpoint for fusion leads to more information preservation during temporal fusion. This is in contrast with \textit{Early Fusion} and \textit{Late Fusion}, where the temporal features are only learned in the viewpoint of the most recent sweep. They do not account for the large viewpoint changes which leads to more information loss.

\subsection{Multi-Sweep Range View Image Generation}
\label{sec:rv_ts_fusion}

The sequential strategy for temporal fusion in the previous section requires a method to generate a fused RV image from the features of two sequential sweeps. We generate this fused RV image in two steps. First, we compensate for ego-motion by transforming the previous sweep into the coordinate frame of the next sweep. The transformed sweep is then projected into the RV and merged with the other sweep. Specifically, let $\mathcal{S}_{m}$ and $\mathcal{S}_{n}$ with $n > m$ be the sweeps to be merged. We first transform $\mathcal{S}_{m}$ into the coordinate frame of $\mathcal{P}_{n}$. Then we use $S_{m,n}$ and $S_{n}$ to generate the fused RV image. For each cell in the fused image, we determine the points from each sweep that project into it using Eq.~\ref{eq:projection}. For each cell $i$, let $\bm{p}^{i}_{m,n}$ and $\bm{p}^{i}_{n}$ be the points that project into it. Each point has a corresponding vector of features, $\bm{f}^{i}_{m}$ and $\bm{f}^{i}_{n}$, which are either hand-crafted, as described in Section~\ref{sec:input}, or learned by a neural network.
These features describe the surface of an object along the ray emanating from the origin of the coordinate system and passing through $\bm{p}^{i}_{m,n}$ and $\bm{p}^{i}_{n}$.
However, due to motion of objects in the scene and/or ego-motion, the features could be describing different objects. Therefore, we provide the relative displacement between the points in the common viewpoint as an additional feature,
\begin{equation}
    \bm{h}^{i}_{m,n} = \bm{R}_{\theta^{i}_{n}}[\bm{p}^{i}_{m,n} - \bm{p}^{i}_{n}]
\end{equation}
where $\bm{R}$ is a rotation matrix parameterized by the angle $\theta^{i}_{n}$.
The resulting features corresponding to the cell~$i$~are, $\bm{f}^{i} = [\bm{f}^{i}_{n}, \bm{f}^{i}_{m,n}, \bm{h}^{i}_{m,n}]$. Although we have described the fusion for a pair of sweeps, this procedure can be trivially extended to any number of sweeps.

\subsection{Detection and Trajectory Estimation}
\label{sec:targets}
The spatio-temporal features in the RV are used to learn multi-scale features through a backbone network (Fig.~\ref{fig:overview}B). The features for the points $\{\bm{p}^{i}_{0}\}$ in $\mathcal{S}_0$ are computed by indexing the cells in the final layer of the backbone using Eq.~\ref{eq:projection}. For each point $\bm{p}^{i}_{0}$, we predict outputs for two sets of tasks (Fig.~\ref{fig:overview}C). The first task  performs semantic classification of the point cloud by predicting probabilities, $\{\hat{p}^{i}_{1}, ..., \hat{p}^{i}_{C}\}$, over the predefined set of classes $\{1, ..., C\}$. The second task performs object detection and motion forecasting by predicting an encoding of the 2D BEV bounding boxes~\cite{lasernet} for $t = \{0, ..., T\}$. The box at $t=0$ corresponds to the object detection. The predicted encoding consists of dimension ($\hat{w}^{i}, \hat{h}^{i}$), centers ($\hat{x}^{i}_{t}, \hat{y}^{i}_{t}$), and orientations ($cos2 \hat{\theta}^{i}_{t}, sin 2\hat{\theta}^{i}_{t}$). In addition to the box, we also estimate the uncertainty in motion by predicting along track $\log \hat{\sigma}^{i}_{AT, t}$ and cross track uncertainties $\log \hat{\sigma}^{i}_{CT, t}$ shared across all box corners. We calculate bounding boxes $\bm{\hat{b}}^{i}_{t}$ using the predicted encoding as described in~\cite{lasernet}. 

We post-process these per-point predictions (Fig.~\ref{fig:overview}C) to produce the final object detections and their trajectories by extending the method in~\cite{lasernet}. First, we select the points belonging to a particular class by thresholding the semantic probability. Then, we perform instance segmentation of these points by clustering the predicted detection centers ($\hat{x}^{i}_{0}, \hat{y}^{i}_{0}$) using approximate mean shift as proposed in~\cite{lasernet}. This results in an instance $j$ comprising all the points $i$ in the cluster along with their corresponding box trajectories $\bm{\hat{b}}^{i}_{t}$ and uncertainties ($\hat{\sigma}^{i}_{AT, t}, \hat{\sigma}^{i}_{CT, t}$). Next, for each instance $j$, we estimate the current and future bounding boxes $\bm{\hat{b}}^{j}_{t}$ and their corresponding uncertainties ($\hat{\sigma}^{j}_{AT, t}, \hat{\sigma}^{j}_{CT, t}$) by aggregating the box trajectories and uncertainties of each point $i$ in the cluster, as proposed in~\cite{lasernet}. Finally, we perform non-maximum suppression (NMS) using the $t=0$ boxes to remove duplicate instances and their trajectory. 

\subsection{End-to-End Training}
\label{sec:loss}
We jointly train the network to produce point classification, object detection and trajectory estimation. We model the semantic probabilities as a categorical distribution  and train using focal loss~\cite{retinanet} for robustness to high class imbalance. The classification loss, $\mathcal{L}_{\text{cls}}$ is calculated as,
\begin{equation}
\label{probability}
    \mathcal{L}_{\text{cls}} = \frac{1}{N_{0}} \sum_{i=1}^{N_{0}}\sum_{l=1}^{C} - [\Tilde{c}^{i} = l](1 - \hat{p}^{i}_{l})^{\gamma} \log\hat{p}^{i}_{l}
\end{equation}
where $\Tilde{c}^{i}$ is the ground truth class label of the point $\bm{p}^{i}_{0}$, $[.]$ is the indicator function and $\gamma$ is the focusing parameter~\cite{retinanet}.

For training detection and trajectory estimation, we model the loss in two components, along the direction of motion (along-track) and perpendicular to the direction of motion (cross-track). We rotate the predicted and ground truth bounding box such that the \textit{x}-axis is aligned to match the along-track component and \textit{y}-axis is to the cross-track component. We model both spatial dimensions as Laplace distributions with shared scale across all four corners and train the parameters using KL divergence~\cite{lasernet-kl}. The regression loss, $\mathcal{L}_{\text{reg}}$ is calculated as,
\begin{equation}
    \mathcal{L}_{\text{reg}} = \frac{1}{T+1}\sum_{t=0}^{T} \alpha_{t} \, \Bigg[ \sum_{a \in \{AT, CT\}} \beta_{a} \, \mathcal{L}_{a, t} \Bigg]
\end{equation}
where $\mathcal{L}_{a, t}$ is calculated as,
\begin{equation}
 \mathcal{L}_{a, t} =  \frac{1}{4*M}\sum_{j=1}^{M} \sum_{n=1}^{4} \text{KL}(Q(b^{j}_{a,n,t}, \sigma^{j}_{a,t}) \| \hat{Q}(\hat{b}^{j}_{a,n,t}, \hat{\sigma}^{j}_{a,t}))
\end{equation}
Here, $\text{KL}$ is a measure of similarity between the predicted distribution $\hat{Q}$ and the ground truth distribution $Q$. Coordinates of the $n^{th}$ predicted corner in along-track and cross-track direction are denoted as $\hat{b}^{j}_{AT,n,t}$ and $\hat{b}^{j}_{CT,n,t}$ respectively. The number of objects in the scene is denoted by $M$ and $\alpha$, $\beta$ are weight parameters which are set using cross-validation. The total loss ($\mathcal{L}$) for training the network is $\mathcal{L} = \mathcal{L}_{\text{cls}} + \mathcal{L}_{\text{reg}}$.

\begin{figure}[t]
    \centering
    \begin{subfigure}{0.48\textwidth}
        \includegraphics[width=0.49\linewidth]{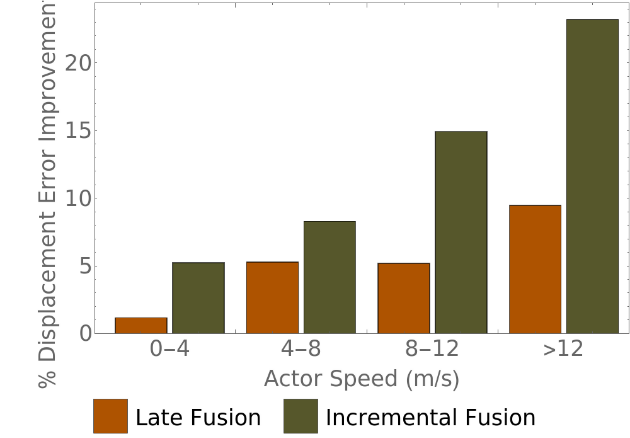}
        \includegraphics[width=0.49 \linewidth]{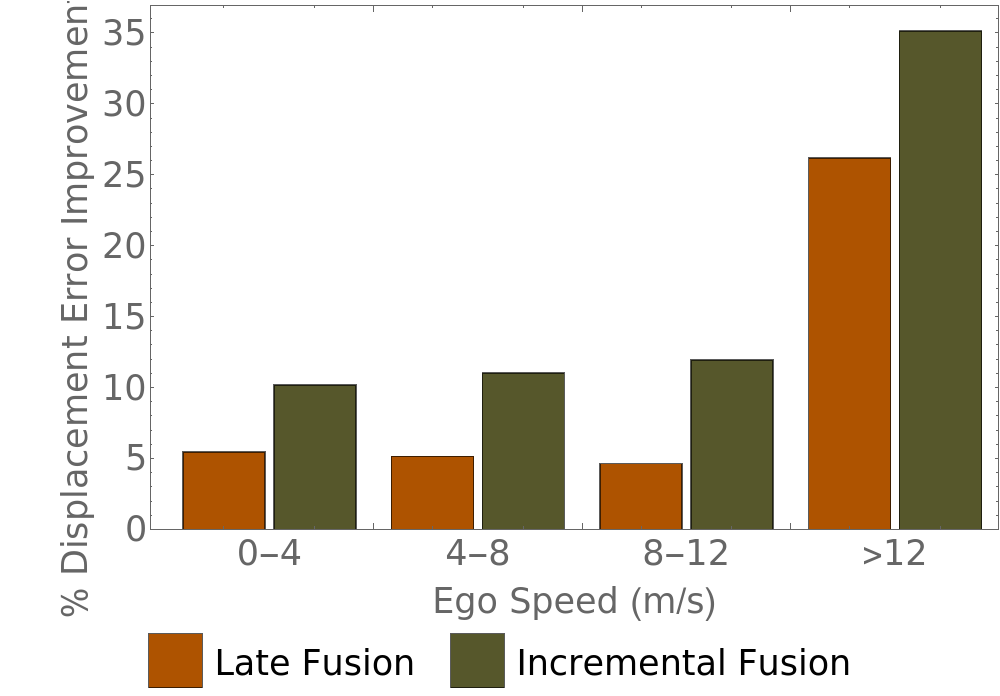}
        \caption{Actor speed and Ego-speed analysis on nuScenes.}
        \label{fig:ablation/nuscenes}
    \end{subfigure}
    \hspace{0.1em}
    \begin{subfigure}{0.48\textwidth}
        \includegraphics[width=0.49\linewidth]{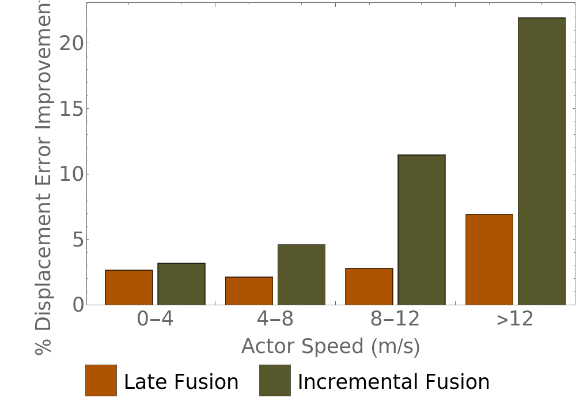}
        \includegraphics[width=0.49 \linewidth]{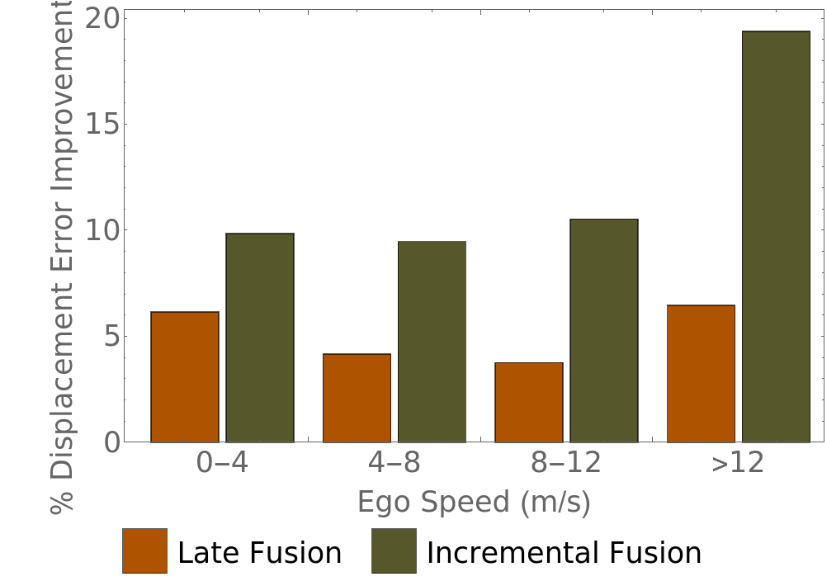}
        \caption{Actor speed and Ego-speed analysis on LSAD.}
        \label{fig:ablation/lsad}
    \end{subfigure}
       \vspace{-0.5em}
    \caption{Plots showing relative improvement (\%) of \textit{Late Fusion} and \textit{Incremental Fusion} over the naive \textit{Early Fusion} on attributes that cause loss of data and structure in the RV. \ref{fig:ablation/nuscenes} and \ref{fig:ablation/lsad} show the improvements on the nuScenes and LSAD datasets, respectively, for different values of actor speed and ego-vehicle. (Sec.~\ref{sec:multi-sweep-fusion-comp} for details)}
    \label{fig:ablation}
\end{figure}

\section{Experiments}
\label{sec:experiments}

\subsection{Dataset and Metrics}
\label{sec:datasets}

We use the publicly available nuScenes~\cite{nuscenes} dataset and an internal Large Scale Autonomous Driving (LSAD) dataset for showing the efficacy of our proposed approach. As compared to nuScenes, our internal dataset has a 4x higher resolution LiDAR and has 5x more data. On nuScenes we train and test on the official region of interest of a square of $100$m, and on LSAD we use a square of $200$m length centered on the ego-vehicle. Following previous work~\cite{spagnn,laserflow}, we train and test on the vehicle objects which includes many fine grained sub-classes such as bus, car, truck, trailer, construction and emergency vehicles.

Following the previous work~\cite{spagnn,laserflow}, we use average precision (AP) for evaluating object detection and $L_{2}$ displacement error at multiple time horizons to evaluate motion forecasting. We associate a detection to a ground truth box if the IoU between them is $\ge \epsilon$ at $t=0$. We compute $L_{2}$ as the Euclidean distance between the center of the predicted true positive box and the associated ground truth box. Like previous methods~\cite{spagnn} \cite{laserflow}, we use $\epsilon = 0.5$ for $L_{2}$ error and $\epsilon = 0.7$ for detections.

\subsection{Implementation Details}
 The size of the range view image is chosen to be $32 \times 1024$ for nuScenes and $64 \times 2048$ for LSAD, based on the sensor resolution. To balance runtime and accuracy, we predict the trajectory for $3$ seconds into the future sampled at $2$Hz, using the LiDAR data from the past $0.5$ seconds sampled at $10$Hz. The implementation is done using the PyTorch~\cite{pytorch} library.

For training, we use a batch size of $64$ distributed over $32$ GPUs. We train for $150$k iterations using an exponential learning rate schedule with a starting rate of $\scnum{2e-3}$ and an end rate of $\scnum{2e-5}$ which decays every $250$ iterations. In the loss we use $\alpha_{0} = 1$ and $\alpha_{t} = 4$ for all $t>0$, $\beta_{AT} = 2$ and $\beta_{CT} = 1$. Further, we use data augmentation to supplement the nuScenes training dataset. Specifically, we generate labels at non-key frames by linearly interpolating the labels at adjacent key frames. We further augment each frame by applying translation ($\pm1$m for $xy$ axes and $\pm0.2$m for $z$ axis) and rotation (between $\pm45^{\circ}$ along $z$ axis) to both the point clouds and labels.

\begin{table}[t]
    \centering
    \scalebox{1.0}{
    \begin{tabular}{l|c|ccc}
        \hline
        \multirow{2}{*}{Method} & \multirow{2}{*}{$AP_{0.7}$ (\%)} & \multicolumn{3}{c}{$L_{2}$ Error (in cm)}\\ \cline{3-5}
        & & 0s & 1s & 3s\\
        \hline
        SpAGNN~\cite{spagnn} & - & \textbf{22} & 58 & 145 \\
        Laserflow~\cite{laserflow} & 56.1 & 25 & 52 & 143 \\
        RV-FuseNet (Ours) & \textbf{59.9} & 24 & \textbf{43} & \textbf{120} \\
    \end{tabular}
    }
    \vspace{-1em}
    \caption{Comparison with state-of-the-art methods on both detection and motion forecasting.}
    \label{tab:nuscenes_sota}
\vspace{-1em}
\end{table}

 \begin{table}[t]
    \centering
    \scalebox{1.0}{
    \begin{tabular}{l|c|ccc}
        \hline
        \multirow{2}{*}{Method} & \multirow{2}{*}{$AP_{0.7}$ (\%)} & \multicolumn{3}{c}{$L_{2}$ Error (in cm)}\\ \cline{3-5}
        & & 0s & 1s & 3s\\
        \hline
        \multicolumn{5}{c}{nuScenes}\\
        \hline
        Early Fusion & 57.9 & \textbf{24} & 48 & 135 \\
        Late Fusion & 58.8 & \textbf{24} & 45 & 127 \\
        Incremental Fusion (Ours) & \textbf{59.9} & \textbf{24} & \textbf{43} & \textbf{120} \\
        \hline
        \multicolumn{5}{c}{LSAD}\\
        \hline
        Early Fusion & 74.1 & \textbf{24} & 41 & 121 \\
        Late Fusion & \textbf{74.6} & \textbf{24} & 40 & 115 \\
        Incremental Fusion (Ours) & \textbf{74.6} & \textbf{24} & \textbf{37} & \textbf{109} \\
        
    \end{tabular}
    }
    \caption{Comparison of RV temporal fusion strategies. Both \textit{Early} and \textit{Late Fusion} perform temporal fusion in the viewpoint of most recent sweep and perform worse than the proposed \textit{Incremental Fusion}.}
    \label{tab:ablation}
\vspace{-1.5em}
\end{table}

\begin{figure*}[t]
    \centering
    \includegraphics[width=0.7\textwidth]{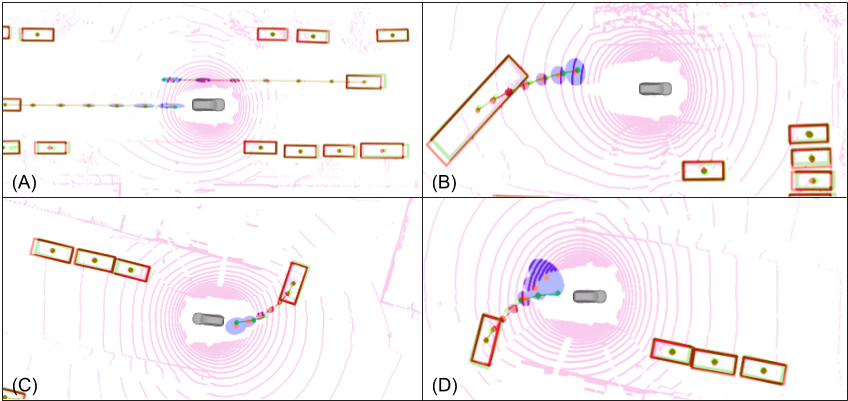}

  \caption{Qualitative examples from the model (red) compared to ground truth (green), with the single standard deviation position uncertainty being plotted in blue. In (A), we can see the model predicting accurate trajectories for straight moving vehicles with tight cross track uncertainty. (B) and (C) show model performance on cases where the actor is turning and has larger uncertainty in the cross-track direction. (D) shows the model failing to accurately predict a turning actor at an intersection, but the true trajectory remains within a single standard deviation of the predicted uncertainty.}
\label{fig:qualitative_results}
\vspace{-1em}
\end{figure*}

\subsection{Comparison with State of the Art}
\label{sec:sota}
In this section, we compare RV-FuseNet to existing methods state-of-the-art joint detection and motion forecasting methods: SpAGNN~\cite{spagnn} and Laserflow~\cite{laserflow}. SpAGNN uses the BEV and Laserflow uses the RV representation of LiDAR for this task. We compare against the publicly available results in their respective publications. As we can see from Table~\ref{tab:nuscenes_sota}, the proposed method clearly outperforms both of these methods by a significant margin and achieves a new state-of-the-art among the joint methods. Particularly, at the $3$s horizon it performs $\sim17\%$ better than both of them. This demonstrates that our proposed method can effectively utilize the high resolution information present in the RV while being robust to the information loss during temporal fusion. Note that we focus on motion forecasting in experimentation for this work. We refer the reader to~\cite{lasernet} for in-depth analysis on object detection, since our approach to detection closely follows it. 

Additionally, Fig. \ref{fig:qualitative_results} shows some real-world examples of our method on the nuScenes dataset, showcasing the ability of our model qualitatively.

\subsection{Comparison of Temporal Fusion Strategies}
\label{sec:multi-sweep-fusion-comp}
In this section, we compare different strategies for temporal fusion of LiDAR data using the RV. For a fair comparison, we use the same multi-scale feature extraction and aggregation backbone (Fig.~\ref{fig:overview}), and the same length of historical LiDAR data for all of these strategies.

In particular, we compare our proposed \textit{Incremental Fusion}, with the naive \textit{Early Fusion} strategy and the \textit{Late Fusion} strategy.  Table~\ref{tab:ablation} details the results on both the nuScenes and LSAD datasets. \textit{Incremental Fusion} provides $\sim10\%$ improvement over the the \textit{Early Fusion} on $L_{2}$ at $3$s as compared to the $\sim5\%$ improvement achieved using \textit{Late Fusion}. This validates the effectiveness of our novel incremental architecture.

When warping the sweep features into a common frame, the amount of information loss and gaps increase as either ego or actor speed increases. In Fig.~\ref{fig:ablation}, we analyze the relative improvements of \textit{Late Fusion} and \textit{Incremental Fusion} over \textit{Early Fusion} in such scenarios. We observe that both \textit{Late Fusion} and \textit{Incremental Fusion} can improve upon \textit{Early Fusion}, showing the benefit of processing each sweep in the original viewpoint before losing information due to the change in viewpoint. Furthermore, we see that in both datasets, \textit{Incremental Fusion} outperforms \textit{Late Fusion} across all speeds. Additionally, we also see that the difference between \textit{Late Fusion} and \textit{Incremental Fusion} dramatically increases as the ego and actor speed increase. This shows the benefit of sequentially changing the viewpoint for fusion of sweeps as compared to fusing all the sweeps in the same viewpoint.

Finally, the consistent improvement of our proposed \textit{Incremental Fusion} over \textit{Late Fusion} and \textit{Early Fusion} across both nuScenes and LSAD shows that it is effective across different resolution LiDARs and dataset sizes. 

\section{Conclusion}
\label{sec:conclusion}

In this work we presented a novel method for joint object detection and motion forecasting directly from time-series LiDAR data using a range view representation. We exploit the high resolution information in the native RV to achieve state-of-the-art results for motion forecasting for self-driving systems. We discuss the challenges of temporal fusion in the range view and propose a novel \textit{Incremental Fusion} model to minimize information loss in a principled manner. Further, we provide an analysis of our method on scenarios prone to information loss and prove the efficacy of our method in handling such situations over alternative fusion approaches.

\clearpage
\bibliography{root}
\bibliographystyle{IEEEtran}
\end{document}